\documentclass[conference]{IEEEtran}
\IEEEoverridecommandlockouts
\usepackage{cite}
\usepackage{amsmath,amssymb,amsfonts}
\usepackage{algorithmic}
\usepackage{graphicx}
\usepackage{textcomp}
\usepackage{xcolor}
\usepackage{tabularx}
\def\BibTeX{{\rm B\kern-.05em{\sc i\kern-.025em b}\kern-.08em
    T\kern-.1667em\lower.7ex\hbox{E}\kern-.125emX}}
\begin{document}

\title{AMF: Adaptable Weighting Fusion with Multiple Fine-tuning for Image Classification
}



\author{
  \IEEEauthorblockN{
    \parbox{\linewidth}{\centering
      Xuyang Shen\IEEEauthorrefmark{1},
      Jo Plested\IEEEauthorrefmark{3},
      Sabrina Caldwell\IEEEauthorrefmark{4},
      Yiran Zhong\IEEEauthorrefmark{1}\IEEEauthorrefmark{2} and
      Tom Gedeon\IEEEauthorrefmark{5}
    }
  }
  \\[-1.5ex]
  \IEEEauthorblockA{
    SenseTime Research\IEEEauthorrefmark{1}, 
    Shanghai AI Lab\IEEEauthorrefmark{2}, 
    University of New South Wales\IEEEauthorrefmark{3}, \\ 
    The Australian National University\IEEEauthorrefmark{4} and
    Curtin University\IEEEauthorrefmark{5}
  }
}

\maketitle

\begin{abstract}
Fine-tuning is widely applied in image classification tasks as a transfer learning approach. It re-uses the knowledge from a source task to learn and obtain a high performance in target tasks. Fine-tuning is able to alleviate the challenge of insufficient training data and expensive labelling of new data. However, standard fine-tuning has limited performance in complex data distributions. To address this issue, we propose the Adaptable Multi-tuning method, which adaptively determines each data sample's fine-tuning strategy. In this framework, multiple fine-tuning settings and one policy network are defined. The policy network in Adaptable Multi-tuning can dynamically adjust to an optimal weighting to feed different samples into models that are trained using different fine-tuning strategies. Our method outperforms the standard fine-tuning approach by 1.69~\%, 2.79~\% on the datasets FGVC-Aircraft, and Describable Texture, yielding comparable performance on the datasets Stanford Cars, CIFAR-10, and Fashion-MNIST.\footnote
{
   
\begin{itemize}
\item    Our code is available:\\ https://github.com/XuyangSHEN/AMF-Adaptable-Weighting-Fusion-with-Multiple-Fine-tuning-for-Image-Classification
\item    Corresponding author: Jo Plested (j.plested@adfa.edu.au)

\end{itemize}
}
\end{abstract}

\begin{IEEEkeywords}
Deep Transfer Learning, Image Classification
\end{IEEEkeywords}

\section{Introduction}
Convolutional neural networks have proved their ability in image classification~\cite{he2016deep,szegedy2017inception,tan2019efficientnet}. In practice, however, there are numerous problems in training accurate convolution neural networks in real-world scenarios. Adequate training data is hard to achieve, and labelling by experts is an expensive task. Reusing a pre-trained model on the existing dataset despite apparent changes in the feature space can lead to poor performance. Furthermore, retraining an extensive model from scratch in a new scenario is also computationally expensive and time-consuming. As a result, transfer learning (i.e. knowledge transfer) is a desirable learning strategy to deal with the above dilemmas~\cite{yosinski2014transferable,pan2009survey}. The amounts of labelled data and computation resources are reduced by transferring the knowledge learned from source domains and tasks into target domains and tasks. Fine-tuning is one common approach in transfer learning, and is the key topic in this research. In fine-tuning, weights of convolutional blocks that are learned from the source task and domains are transferred and retrained on new domains and tasks. These trained weights are proved to out-perform random-sampled weights in many new domains, which demonstrates that fine-tuning is a practical approach~\cite{jarrett2009best,girshick2014rich,azizpour2015factors,mahajan2018exploring,plested2019analysis,kornblith2019better}. 

We focus on mixture distributions, being probability distributions that are controlled by two or more modes, in this research. The mixture distribution is a typical distribution in the real world, but it is difficult for standard neural networks to implement with good performance in accuracy and efficiency~\cite{gao2020survey}. Our empirical results indicate that the performance of standard fine-tuning is restricted in complex mixture distributions. As far as we know, none of the existing research in transfer learning targets complex data distributions. Our research explores fine-tuning approaches on complex data distributions.

\begin{figure*}[t]
      \centering
      \includegraphics[width=\textwidth]{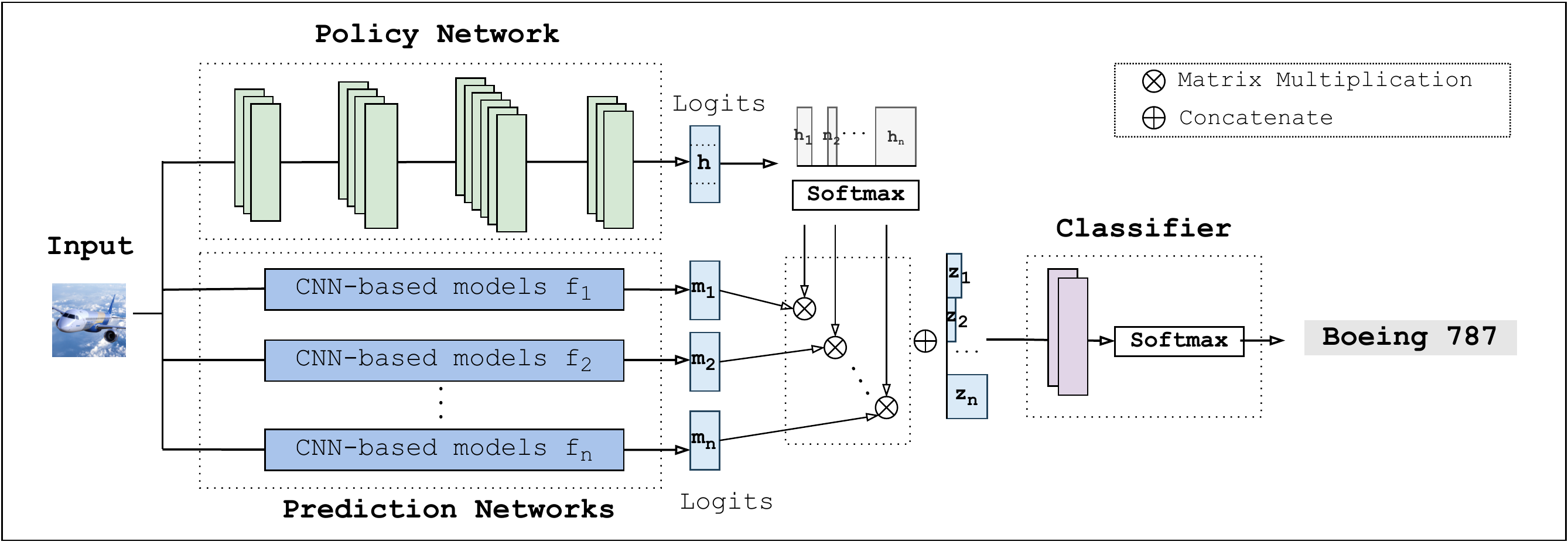}
      \caption{\textbf{Adaptable Multi-tuning Framework}}
      \label{fig:amt}
\end{figure*}

In this paper, we propose Adaptable Multi-tuning Framework (AMF) as shown in Figure~\ref{fig:amt}, an approach that can adaptively weight features from multiple fine-tuning strategies in the final classifier through a policy network. The pipeline of AMF is inspired from SpotTune which can dynamically adjust each layer of the network from freezing and fine-tuning~\cite{guo2019spottune}. Our approach consists of a prediction module and policy network. The prediction module are constructed from more than one fine-tuned models. The policy network learns the weight of feature spaces which are calculated from each sub-model in the prediction module. In order to make the decision policy differentiable during training, we transfer the decision into continuous values and apply $Softmax$ to unify the values. A fully-connected layer as classifiers is the last components of Adaptable Multi-tuning Framework. These accept the weighted latent space derived from the prediction module and policy network and outputs classification results.

The contributions of our papers are as follows: (1) We construct two challenge datasets in mixture distribution: Aircraft-DTD and Aircraft-Cars, from FGVC-Aircraft~\cite{maji2013fine}, Stanford Cars~\cite{krause20133d}, and Describable Textures Dataset~\cite{cimpoi2014describing}. (2) We show that standard fine-tuning has good performance on simple mixture distribution, but is less accurate in the complex distribution domain. (3) We propose Adaptable Multi-tuning Framework for strong and consistent performance on both complex and straightforward mixture distribution. It overcomes the limited performance of standard fine-tuning.

\section{Related Work}

\subsection{Deep Transfer Learning in Image Classification}

Deep transfer learning has been proved to achieve comparable accuracy of image classification~\cite{jarrett2009best,girshick2014rich,yosinski2014transferable,azizpour2015factors,mahajan2018exploring,plested2019analysis,kornblith2019better,he2019rethinking,neyshabur2020being,plested2021non}. Freezing pre-trained weights of convolutional neural networks from a large dataset as a feature extractor, and training a simple classification model for target tasks, is common in transfer learning~\cite{azizpour2015factors,donahue2014decaf,sharif2014cnn}. The transferred feature extractor can outperform the models trained from scratch under positive transfer; even when the target task is not highly correlated to the source one. Fine-tuning is another common approach for transfer learning. Empirical results also show that fine-tuned weights usually out-perform frozen weights~\cite{yosinski2014transferable,girshick2014rich,azizpour2015factors,huh2016makes,chu2016best,mahajan2018exploring,kornblith2019better,plested2019analysis}. Recent studies also indicate that learning rates, momentum value and decay values significantly control the impact of pre-trained weights to training on the target dataset~\cite{sun2017revisiting,li2020rethinking,plested2021rethinking}. In line with these discoveries in~\cite{yosinski2014transferable}, a low learning rate is suitable for low-level convolutional blocks, and relatively high value settings are suitable for high-level convolutional blocks. lower learning rates and higher decay rates are preferred for fine-tuning in more similar source and target tasks~\cite{kolesnikov2020big,li2020rethinking,plested2021non}. Additionally, fine-tuning more layers is beneficial for increasing the performance when source and target datasets are more related~\cite{yosinski2014transferable,chu2016best,li2020rethinking,plested2021rethinking}. 

Transfer learning is still useful when the target domain is already sufficient for models to train from scratch\cite{yosinski2014transferable,sun2017revisiting,mahajan2018exploring}. On the other hand, fine-tuning cannot prevent over-fitting and biased empirical loss estimation in small source and target datasets~\cite{he2019rethinking,kolesnikov2020big,mahajan2018exploring}. For this reason, ImageNet is a common choice for the source dataset. Nevertheless, fine-tuning on closely related datasets can obtain better performance~\cite{mahajan2018exploring,plested2021rethinking}

\subsection{Transferability}

Transferability is a recessive factor influencing the desirable behaviour (i.e. transfer's feasibility) of deep neural networks that learn from other tasks or domains. Features from lower layers of deep neural networks are more general, regardless of the domain and tasks~\cite{yosinski2014transferable,plested2019analysis,he2019rethinking,plested2021non}; thus they have high transferability. Features become more specific, discrete and differential when layers become deeper~\cite{yosinski2014transferable,liu2019towards,plested2019analysis,he2019rethinking,plested2021non}. H-score, conditional entropy, and multi-class Fisher score are common methods for measuring transferability~\cite{bao2019information,tran2019transferability,plested2021rethinking}. Transferability can not only provide an insightful explanation of the transfer learning mechanism, but can also measure the performance of pre-trained models in target dataset and guide the hyper-parameters tuning~\cite{bottou2007tradeoffs,bao2019information,tran2019transferability}. We illustrate the transferability of our proposed method by feature analysis in analysis section. 

\section{Proposed Approach}

\subsection{Multiple Fine-tuning}
A mixture distribution is a complex probability distribution. It refers to a probability distribution with two or more modes. The bimodal distribution is a typical distribution from a mixture distribution, but it is difficult for deep neural networks to obtain a good performance in both accuracy and efficiency with a small training dataset~\cite{gao2020survey,sohn2014improved}. Our experiments indicate that standard fine-tuning struggles to achieve competent performance in mixture distributions, because of limited transferability. Based on this motivation, we propose a new idea, adaptable multi-tuning, which can improve the performance on mixture distribution datasets and still achieve performance comparable to SoTA on normal datasets. 

Standard fine-tuning achieves good performance on standard classification; hence, we aim to stack several different fine-tuned models in parallel as a complex feature extractor and combine all hidden features identically to construct the classifier. A similar architecture has been proposed in MultiTune~\cite{wang2020multitune} and mixture of experts model~\cite{shazeer2017outrageously,ramachandran2018diversity,crawshaw2020multi,hazimeh2021dselect}. MultiTune uses pre-defined weight fusion, and slightly outperforming SpotTune and standard fine-tuning. We target on an adaptable fusion mechanism that is able to flexibly adjust the contribution of each fine-tuned model to the final classifier.


\begin{figure}[t]
      \centering
      \includegraphics[width=\columnwidth]{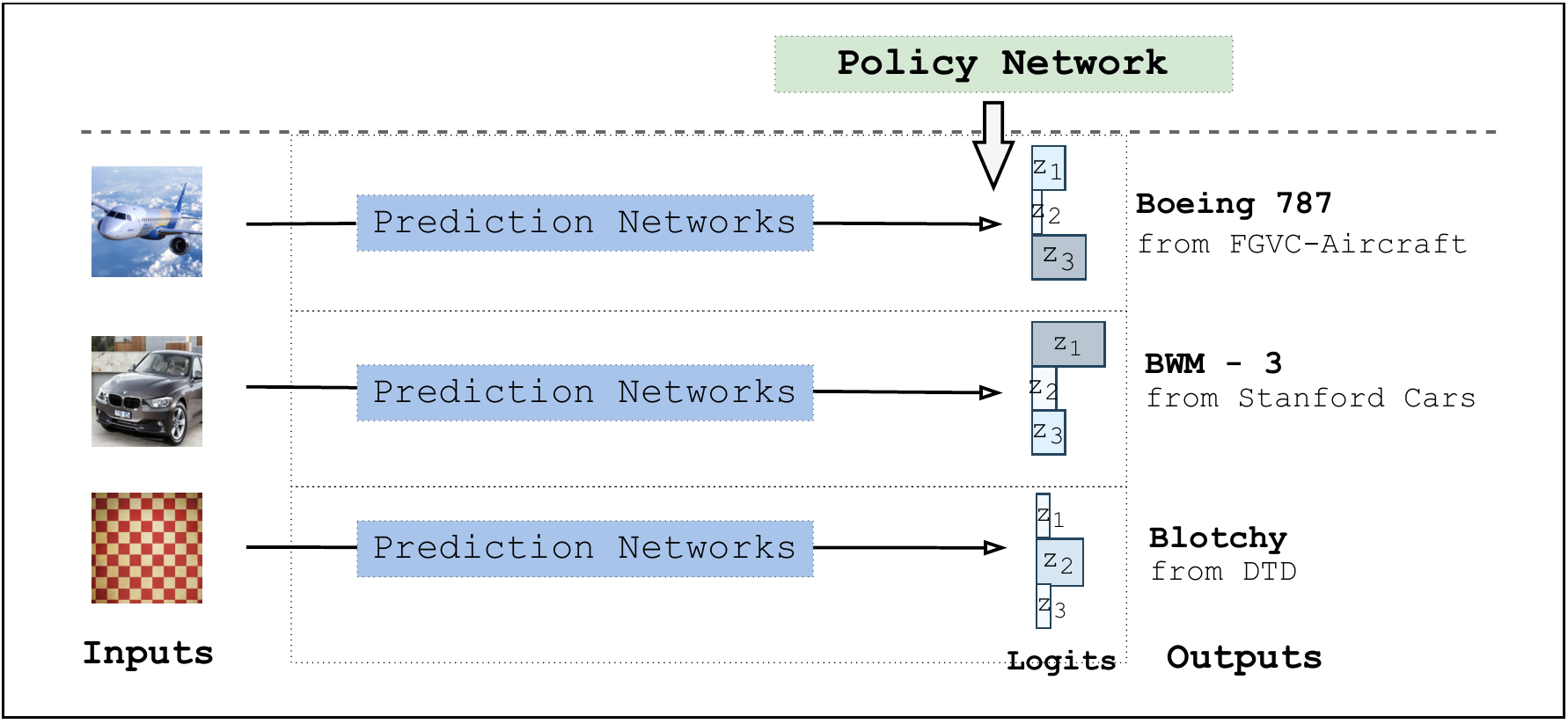}
      \caption{\textbf{The Pipeline of Adaptable Multi-tuning Framework}. The rectangle with deeper color and longer width indicates the average value of prediction from policy network is closer to $1$; the weighted latent vector has more potential to predict the correct class.}
      \label{fig:amt_samples}
\end{figure}

\subsection{Adaptable Fusion}

To achieve adaptable adjustment based on the input distribution, we design the policy network. Given the current data samples, it produces a weights vector. The classifier can regress the outputs based on the scaled feature matrix. Loss is calculated to back propagate the policy network, directly impacting the weights of policy network that cause errors. Additionally, $Softmax$ is used to normalize the output of the policy network.

\subsection{Adaptable Multi-tuning}
Figure~\ref{fig:amt} illustrates the overall structure of adaptable multi-tuning. It consists of three primary sections: a policy network, a prediction network, and a classifier. The prediction network acts as the feature extractor, constructed by several fine-tuning settings. Each fine-tuned model ($F_{i :\ i\in [1,n]}$) accepts inputs in parallel from a mixture distribution target dataset, and outputs a fixed-length matrix  ($m_{i :\ i\in [1,n]}$) as latent spaces. The inputs are also passed into the policy network ($F_{policy}$) and softmax operation ($\sigma$), producing a weight vector whose shape is $1 \times n$. The weighted latent space ($z_{i :\ i\in [1,n]}$)  is calculated from the scaled policy output ($h_i$) and latent space ($m_{i :\ i\in [1,n]}$) through matrix multiplication and concatenation. The policy network learns to scale the essential features from a group of features derived from the prediction network through these operations. If an ideal performance is reached, the policy network is able to predict a value close to zero for the latent space that does not beneficial for prediction networks to perform classifications; on the other hand, it can indicate a value close to one. It is functionally equivalent to searching for the most suitable fine-tuning settings for each example from the input distribution. The scale vector ($h_i$) outputted by the policy network is considered as confidence scores, based on its functionality. Figure~\ref{fig:amt_samples} plots three samples of weighted latent spaces for data from different distributions. 

The classifier of adaptable multi-tuning is a standard fully-connected layer with softmax function. It accepts the concatenation of weighted latent spaces ($z_{i :\ i\in [1,n]}$) as input. The concatenation fusion is chosen as it can preserve the numerical meaning better, compared to other operations such as addition.

Adaptable Multi-tuning is designed to accept any value of hyper-parameters $n$ (number of fine-tuning settings), and is determined by the input distribution. For example, $n=2$ is recommended for inputs from a bimodal distribution. In addition, an example of $n=3$ is shown in Figure~\ref{fig:amt_samples}. Furthermore, the design intent for policy networks is not to increase the computation stress by an additional network module. For this reason, for the network architecture of policy networks, we prefer it to be simpler and more shallow than models in the prediction network, such as default ResNet-34. 

In standard fine-tuning, it is impossible to coordinate the learning rates with various classes in mixture distributions. In contrast, Adaptable Multi-tuning with prediction module and policy network is expected to have a good performance in mixture distributional classification tasks.  The prediction module is designed to accept multiple learning rates. The policy network is introduced to control the weights for different prediction models across various learning rates.

\section{Experiments}

\subsection{Experimental Setup}

\subsubsection{Dataset}
We select ImageNet-1K~\cite{deng2009imagenet} as the source dataset. Five public target datasets are selected to compare our adaptable multi-tune with traditional fine-tuning, including two fine-grained tasks FGVC-Aircraft~\cite{maji2013fine} and Stanford Cars~\cite{krause20133d}, one dataset with large domain difference to ImageNet-1K: DTD~\cite{cimpoi2014describing}, and two other datasets: CIFAR-10~\cite{krizhevsky2009learning}, and Fashion-MNIST~\cite{xiao2017fashion}.

\subsubsection{Mixture Distribution Setup}
\label{misture_distribution_setup}

\begin{table}[tb]
\caption{\textbf{Statistics Summary of Mixture Distribution Setup}. Columns in italic format are used to distinguish the \textbf{O}riginal dataset from \textbf{M}odified dataset. First $47$ classes of Aircraft are selected to construct Aircraft-47; Cars-75 are built from the first $75$ classes of Cars; First $33$ samples in each class of DTD are assembled to DTD-47.}
\centering
\label{tab:ssmpd}
\begin{tabular}{lllll}
\hline
                     & \textit{O. classes} & \textit{O. size} & M. classes & M. size \\ \hline
\textbf{Aircraft-47} & \textit{100}              & \textit{3333}          & 47               & 1551          \\ \hline
\textbf{Cars-75}     & \textit{196}              & \textit{4616-8041}     & 64               & 1721-3032     \\ \hline
\textbf{DTD-47}      & \textit{47}               & \textit{1800}          & 47               & 1551          \\ \hline
\end{tabular}
\end{table}

\vspace{10mm}

Obtaining the performance of mixture distribution is the main target; thus, two of FGVC-Aircraft, Stanford Cars, and DTD are grouped to construct complex data distributions (i.e. bi-model distributions). CIFAR-10 and Fashion-MNIST are existing mixture distribution with around  $60,000$ training examples; thus no further modification are performed. In order to minimize the classes biases by coordinating the total size of each component into a uniform number, both number of classes and number per class is adjusted as shown in Table~\ref{tab:ssmpd}. Specifically, the number of Aircraft classes is reduced to $47$ (Aircraft-47); number of Cars classes is reduced to $75$ (Cars-75); number per class of DTD is reduced to $33$ (DTD-47).

\begin{figure}[tb]
      \centering
      \includegraphics[width=0.9\columnwidth]{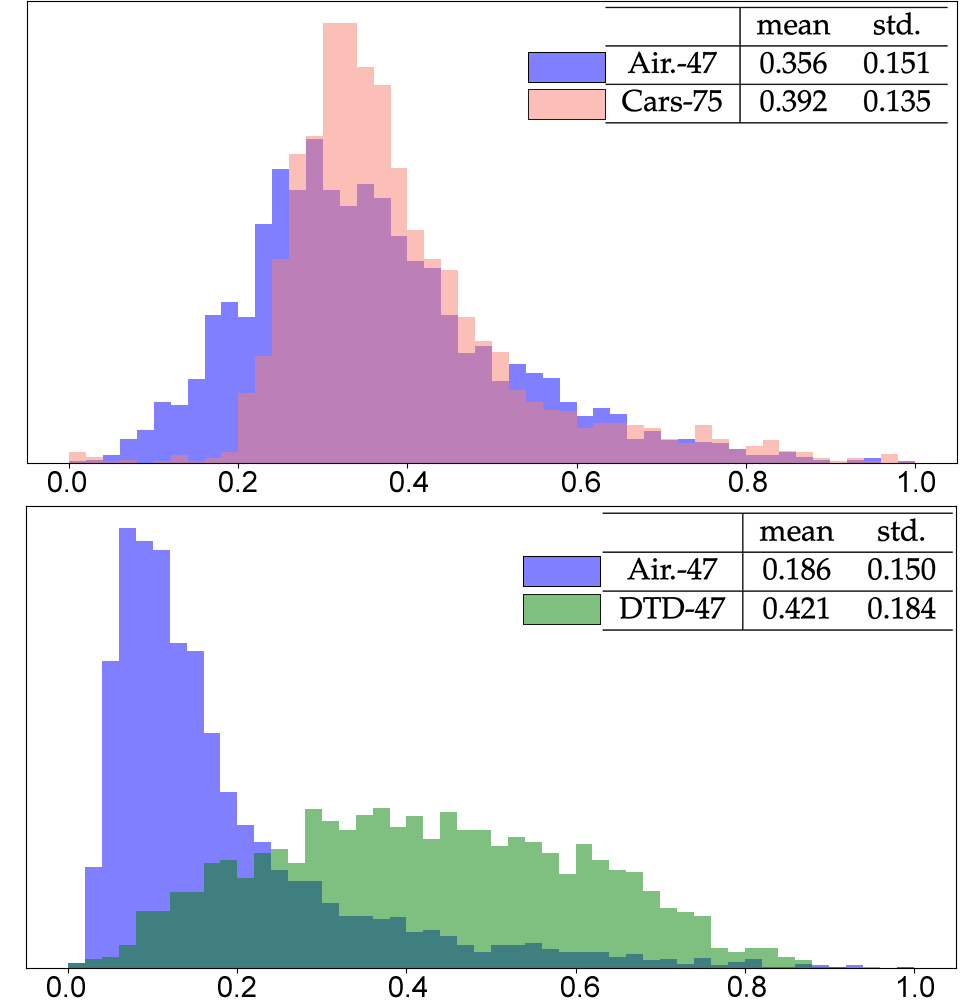}
      \caption{ \textbf{T-SNE~\cite{hinton2002stochastic} Visualisation (in one-dimension) of Feature Distribution in Aircraft-Cars (left) and Aircraft-DTD (right)} }
      \label{fig:airdtdcar_dist}
\end{figure}

Aircraft and Cars are fine-grained tasks, and DTD has a large difference in terms of visual intuition to ImageNet-1K~\cite{guo2019spottune,plested2021rethinking}. Therefore, we decided to group Aircraft-47 and Cars-75 as one complex distribution dataset; and Aircraft47 and DTD-47 as another complex distribution dataset. We use well-trained models as feature extractors to analyze the features of the above mixture-distribution datasets as shown in histograms~\ref{fig:airdtdcar_dist}. The two sub-groups in Aircraft-Cars have closer mean values and overlapping peaks, while the two sub-groups in Aircraft-DTD have two distributed peaks. These combined datasets can evaluate different methods from two perspectives: a complex distribution with similar features, and a complex distribution with different features. A technical summary of the two new datasets:
\begin{itemize}
\item \textbf{Aircraft-DTD} contains $94$ classes and $33$ images per class in train, test, and validate. Half the classes are from Aircraft-47, and the other are from DTD-47. 
\item \textbf{Aircraft-Cars} consists of $3102$ and $3071$ train-validation samples from aircraft and cars, respectively. There are $33$ aircraft classes and $75$ car classes in the mixture dataset. 
\end{itemize}

\subsubsection{Backbone}

In recent work, Inception-v4 with novel fine-tuning strategy~\cite{plested2021rethinking} achieved several state-of-the-art results on the above datasets. To illustrate a comparable result, Inception-v4 is selected for our experiments as the backbone~\cite{szegedy2017inception}.

\subsubsection{Metrics}
Top-1 accuracy is used to evaluate all experiments, since it is the conventional performance measure in image classification tasks. Note that final accuracy is averaged over five runs.

\subsubsection{Model}

Based on the data distributions and backbone, a specific Adaptable Multi-tuning Network from the general architecture (Figure~\ref{fig:amt}) is designed. This model accepts two pre-defined fine-tuning hyper-parameters for two Inception-v4s. ResNet-34 with four blocks $(3,4,6,3)$ is used in the policy network. Inception-v4 and ResNet-34 are separately pre-trained on ImageNet-1K. We use cross-entropy as the loss function in entire experiments.


\subsubsection{Experiments}
We compare our proposed adaptable multi-tune with the following methods:
\begin{itemize}
\item \textbf{Standard Fine-tune}: All pre-trained weights of Inception-v4 are fine-tuned on the target dataset.  

\item \textbf{MultiTune: a static multiple fine-tune~\cite{wang2020multitune}}: We re-implement MultiTune with Inception-v4. It consists of two backbones in feature extraction, followed by a fixed concatenation fusion operation. All pre-trained weights of the backbones are fine-tuned.
\end{itemize}

In the standard fine-tuning method, three learning rates are set~\cite{plested2021rethinking}, including values for shallow blocks, deep blocks, and final fully-connected layers. MultiTune also requires three learning rates, where one each for sub-networks and one for final fully-connected layers; while an additional learning rate is required for policy network in our proposed method. The same batch size and decay rates are used for the entire network for each method.

\subsubsection{Implementation}

We use the pre-trained Inception-v4 model from~\cite{rw2019timm} as the starting point. All experiments are  run on a single A-100 (32Gb) GPU. We use the original data separations of train, validation and test for the above datasets. Since the split of training and validation are not provided in Stanford Cars, we take first 18 examples in each class as validation. Furthermore, SGD with momentum is used as the optimizer. Random augment and random erasing are also enabled, which is set to $rand-m9-mstd0.5-inc1$ and $0.5$~\cite{cubuk2020randaugment}.

\subsection{Results and Analysis}

\begin{table*}[t]
\caption{\textbf{Baseline Fine-tune (SoTA) on Target Datasets, measured in Top-1 accuracy ($\%$)}. Original pubic dataset stands for public datasets; modified dataset refers to our set-up in Section~\ref{misture_distribution_setup}. Note that, CIFAR-10 and Fashion-MNIST are not modified.}
\centering
\label{tab:base_acc2}
\begin{tabularx}{\textwidth}{@{}l *9{>{\centering\arraybackslash}X}@{}}
\hline
\textbf{Dataset}      & \multicolumn{2}{c}{\textbf{Aircraft}} & \multicolumn{2}{c}{\textbf{Cars} }   & \multicolumn{2}{c}{\textbf{DTD}}    & \textbf{CIFAR} & \textbf{Fmnist} \\ \hline
Split        & Val               & Test              & Val              & Test             & Val              & Test             & Acc               & Acc              \\ \hline
\textbf{Original}     & 89.20  & 94.36  & 92.25 & 94.81 & 71.12 & 77.87 & 98.43  &95.26 \\
\textbf{Modified} & 84.53  & 93.20  & 87.78 & 90.95 & 70.02 & 76.41 & -        & -       \\ \hline
\end{tabularx}
\end{table*}

\begin{table*}[t]
\caption{Top-1 test accuracy ($\%$) of Adaptable Mulit-tune Network (AMF), baselines, and MultiTune on Aircraft-DTD, Aircraft-Cars, CIFAR-10, and Fashion-MNIST. Accuracy difference (STD) among 5 runing times of AMF is also reported. }
\centering
\label{tab:results}
\begin{tabularx}{\textwidth}{@{}l *7{>{\centering\arraybackslash}X}@{}}
\hline
\textbf{Dataset}   & \multicolumn{2}{c}{\textbf{Aircraft-DTD}} & \multicolumn{2}{c}{\textbf{Aircraft-Cars}} & \textbf{CIFAR-10 \ \ } & \textbf{F-MNIST \ \ } \\ \hline
sub-class          & Aircraft-47           & DTD-47            & Aircraft-47            & Cars-75           & -                 & -                      \\ \hline
\textbf{Fine-tune} & 91.32               & 74.12           & 92.09                & 90.59           & 98.43           & 95.26                \\
\textbf{MultiTune~\cite{wang2020multitune}} & 88.55               & 75.91           & 86.69                & 89.50           & 98.47           & 95.27                \\ \hline
\textbf{AMF}       & 93.01               & 76.91           & 92.50                & 90.79           & 98.50           & 95.59                \\ 
\textbf{STD}  & $\pm0.2$  & $\pm0.2$  & $\pm0.4$ & $\pm0.2$ & $\pm0.05$ & $\pm0.08$ \\
\hline
\end{tabularx}
\end{table*}

\subsubsection{Fine-tune Baseline\\}

Empirical results of the standard fine-tuning methods on standard datasets are shown in Table~\ref{tab:base_acc2}. Through tweaking the learning rates in five validation sets to $0.03$, $0.025$, $0.0008$, $0.03$, and $0.025$ (same order in the Table), we reproduce comparable test accuracy to~\cite{plested2021rethinking} on the full-sized target datasets. Learning rates in the first $8$ layers are reduced $60\%$ to preserve prior knowledge learned from ImageNet-1K. 

When the size is reduced in Aircraft, Cars, and DTD, the number of samples in each epoch is reduced, causing accelerated decline in learning rates. Therefore, the optimal hyper-parameters for the full-sized dataset are maintained in the reduced-size dataset, except for decay rates. Empirical accuracy differences from the reduced-size dataset and the original dataset reflects that our modification increases task difficulty as the training datasets are all smaller. 
The performance of standard fine-tuning on the test set of two mixture distribution datasets (i.e. Aircraft-DTD and Aircraft-Cars) is shown in Table~\ref{tab:results}, which is also a comparable baseline to our methods. We start finding hyper-parameters by reusing the settings in the above experiments. In Aircraft-DTD, a high learning rate ($0.025$) leads to the model performing more accurately in Aircraft and less accurately in DTD. On the other hand, when a lower learning rate is set ($0.0008$), the model produces strong performance in DTD, but accuracy of Aircraft drops $10\%$. It is infeasible to coordinate disparate learning rates with two classes in standard fine-tuning, as expected. The final test accuracy of Aircraft-47 and DTD-47 in Aircraft-DTD are $91.32\%$ and $74.12\%$, while $93.20\%$ and $76.41\%$ are the best test accuracy in individual Aircraft-47 and individual DTD-47. The decline in accuracy also happens in Aircraft-Cars for both validation and test accuracy, although both aircraft and cars prefer high learning ($0.025$) for fine-tuning. The final test accuracy of Aircraft-47 and Cars-75 in Aircraft-Cars: $92.09\%$ and $90.59\%$. 
 
By comparing the accuracy between Aircraft-DTD and Aircraft-Cars, the former dataset is difficult for standard fine-tuned models to train and classify. It is reflected by the feature distribution ( Section~\ref{misture_distribution_setup}) of these two datasets, in which aircraft and cars share more similar features, whereas aircraft and describable textures are located more distantly in feature space. Therefore, we can conclude that standard fine-tuning has limited performance when dealing with complex distribution domains.

\subsubsection{Performance of AMF\\}

The results of the comparison between standard fine-tuning, MultiTune~\cite{wang2020multitune}, and Adaptable Multi-tune Network is shown in Table~\ref{tab:results}. Our proposed methods yields consistently better test accuracy than standard fine-tune and multi-tune in all four datasets. In mixture distribution datasets, Adaptable Multi-tuning Network outperforms $1.69\%$ and $2.79\%$ in Aircraft-47 and DTD-47 respectively, than the standard fine-tune on Aircraft-DTD dataset; our method also achieves comparable performance on Aircraft-Cars compared to standard fine-tuning, and outperforms $0.41\%$ and $0.20\%$ in Aircraft-47 and Cars-75 respectively. Adaptable Multi-tuning Network also holds a slight lead in CIFAR-10 and Fashion-MNIST.

It might be unfair to directly compare the performance of our method with standard fine-tuning, since the former uses two Inception-v4 and the latter uses one Inception-v4. The parameters of Adaptable Multi-tuning Networks ($103.6M$) are twice larger than standard fine-tune ($41.2M$) in experiments. Therefore, we also report performance of standard MultiTune ($83.1M$) in Table~\ref{tab:results}. Our method holds a large improvement, compared to the MultiTune in both Aircraft-DTD and Aircraft-Cars. This reflects that the policy network contributes benefit to model performance.

In summary, our Adaptable Multi-tune Network has stable performance for all above datasets. In contrast, the standard fine-tune and MultiTune method only achieves comparable accuracy on CIFAR-10 and Fashion-MNIST.

\subsubsection{Hyper-parameters Optimization of AMF} 

\begin{figure*}[t]
  \centering
  \includegraphics[width=0.9\textwidth]{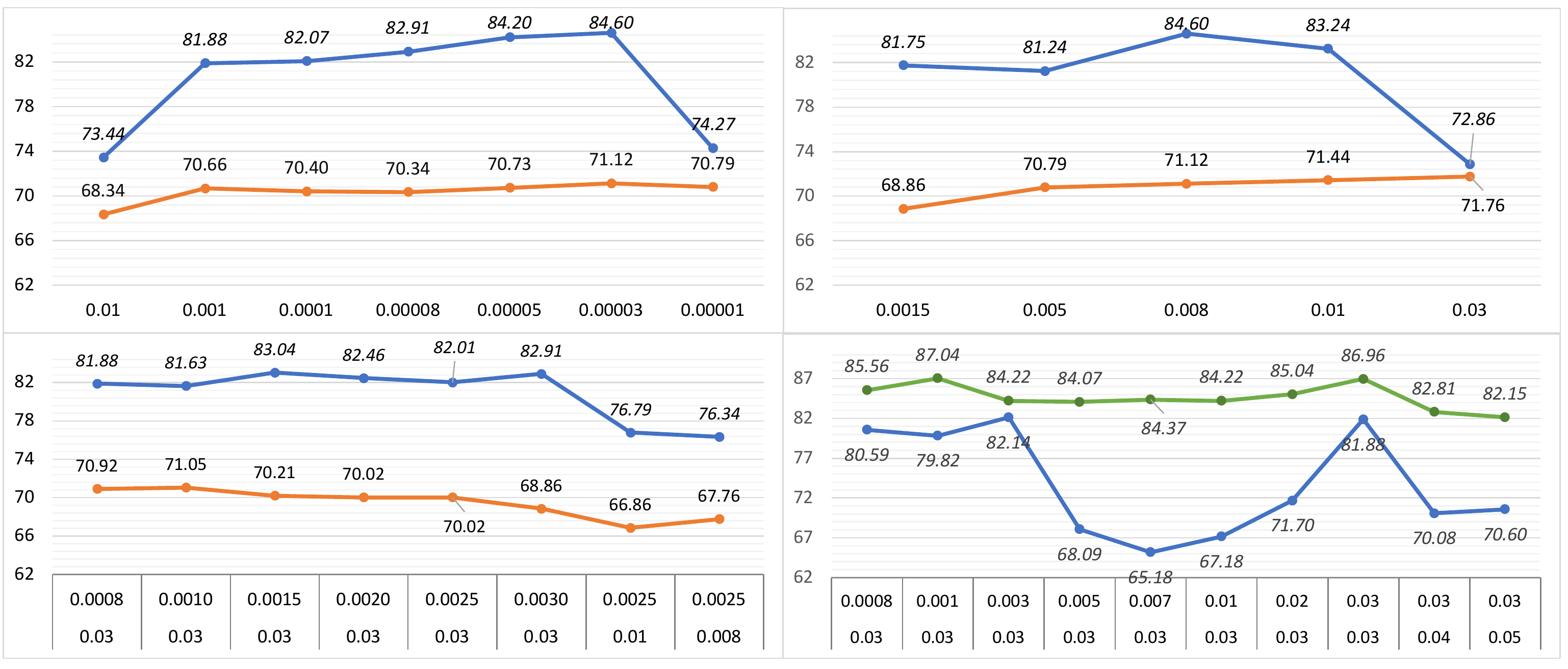}
  \caption{\textbf{Hyper-parameters Search of AMF in Aircraft-DTD and Aircraft-Cars. }Blue: Aircraft-47 validation accuracy; Orange: DTD-47 validation accuracy; Green: Cars-75 validation accuracy. Figures from top-right to bottom-right: learning rates of policy network optimization; learning rates of fully-connected layers optimization; learning rates of Inception-v4 optimization in Aircraft-DTD; learning rates of Inception-V4 optimization in Aircraft-Cars.}
  \label{fig:hps}
\end{figure*}

\textbf{\\Learning rates:} Adaptable Multi-tuning Network requires four pre-defined learning rates. The policy network aims to predict which fine-tuning hyperparameters are most suitable for each individual image sample. To simulate a weak penalty mechanism, we set a tiny learning rate for policy networks to scale down the strong gradient descent from cross-entropy loss in a reasonable range. Additionally, policy networks might suffer a heavy error penalty because of incorrect classification by prediction modules (i.e. Inception-v4). A tiny learning rate is also necessary to minimize this effect. Pre-trained weights for policy networks are also required, to extract features at the initial stage of training. The empirical results shown in the top-left sub-plot~\ref{fig:hps} support our hypothesis that the policy network learning rate should be small. A high learning rate for the policy networks leads to a steep accuracy decline of the aircraft classes in Aircraft-DTD. Learning rates between $0.00001$ and $0.0001$ are suitable for the policy network, and $0.00003$ is the optimal value.

Our hypothesis about the optimal learning rates for the prediction modules is based on the purpose in designing them. Our design idea is that each fine-tuning network is able to learn individual data distributions separately in a mixture-distribution. In other words, the optimal learning rates should be closer to values in standard fine-tuning. We perform grid search to determine the best learning rates for each individual module, and the results are plotted in Figure~\ref{fig:hps}. The model is not sensitive to the various learning rates of fully-connected layers (top-right sub-plot) when it is between $0.005$ and $0.01$. Adjusting the learning rates of fine-tuning modules with low learning rates cannot influence model performance; by contrast, decreasing learning rates of fine-tuning modules with higher learning rates can diminish performance, caused by a dead policy network. We select $0.0008, 0.03, 0.008, 0.00003$ learning rates in our final Aircraft-DTD experiments for sub-network 1, sub-network 2, the final fully connected layer, and the policy network respectively. A similar grid search is also performed in Aircraft-Cars. We conclude that $0.03, 0.03, 0.008, 0.00003$ are the optimal learning rates, with the two sub-networks in the prediction modules having the same learning rates. The optimal learning rates in two mixture distribution datasets are same as the values in standard fine-tuning.

We suggest that (1) optimal learning rates for standard fine-tuning models in individual datasets is also highly possible to be optimal for sub-networks in Adaptable Multi-tuning Network; (2) policy networks perform better with a low learning rate. 

\textbf{Other hyper-parameters:} When decay epochs are set between $20$ and $25$, and the decay rate is set within $[0.9, 0.96]$, Adaptable Multi-tuning Network is slightly affected. On the other hand, predictions of policy network may locate in extreme range. Furthermore, a small batch size ($32$) is preferred in Aircraft-DTD and Aircraft-Cars, while a large batch size ($128$) is suitable in CIFAR-10 and Fashion-MNIST.

\subsection{Insight of AMF}

Policy networks of Adaptable Multi-tune Networks is highly accurate and quickly converged. The large value of embedding (derived by the probability of policy network) indicates which of the two datasets in the mixture distribution is preferred. In the test-set of Aircraft-DTD, the trained policy network only miss-assigns $24$ among $3102$ mixed samples of Aircraft and DTD; additionally, all DTD samples are correctly assigned. The policy network behaves more accurately in the test-set of Aircraft-Cars, achieving close to $100\%$ accuracy in separating the two data distributions. We also implement a dynamic monitor to record changes in the accuracy of assignment from the policy networks throughout training. The policy network always predicts an embedding space that treats all samples from the same classes at the initial stage; it rapidly adjusts the prediction during the first $50$ epochs; policy networks converge after the first $300$ epochs, where accuracy fluctuation reduces and becomes stable.

\begin{figure*}[t]
  \centering
  \includegraphics[width=0.9\textwidth]{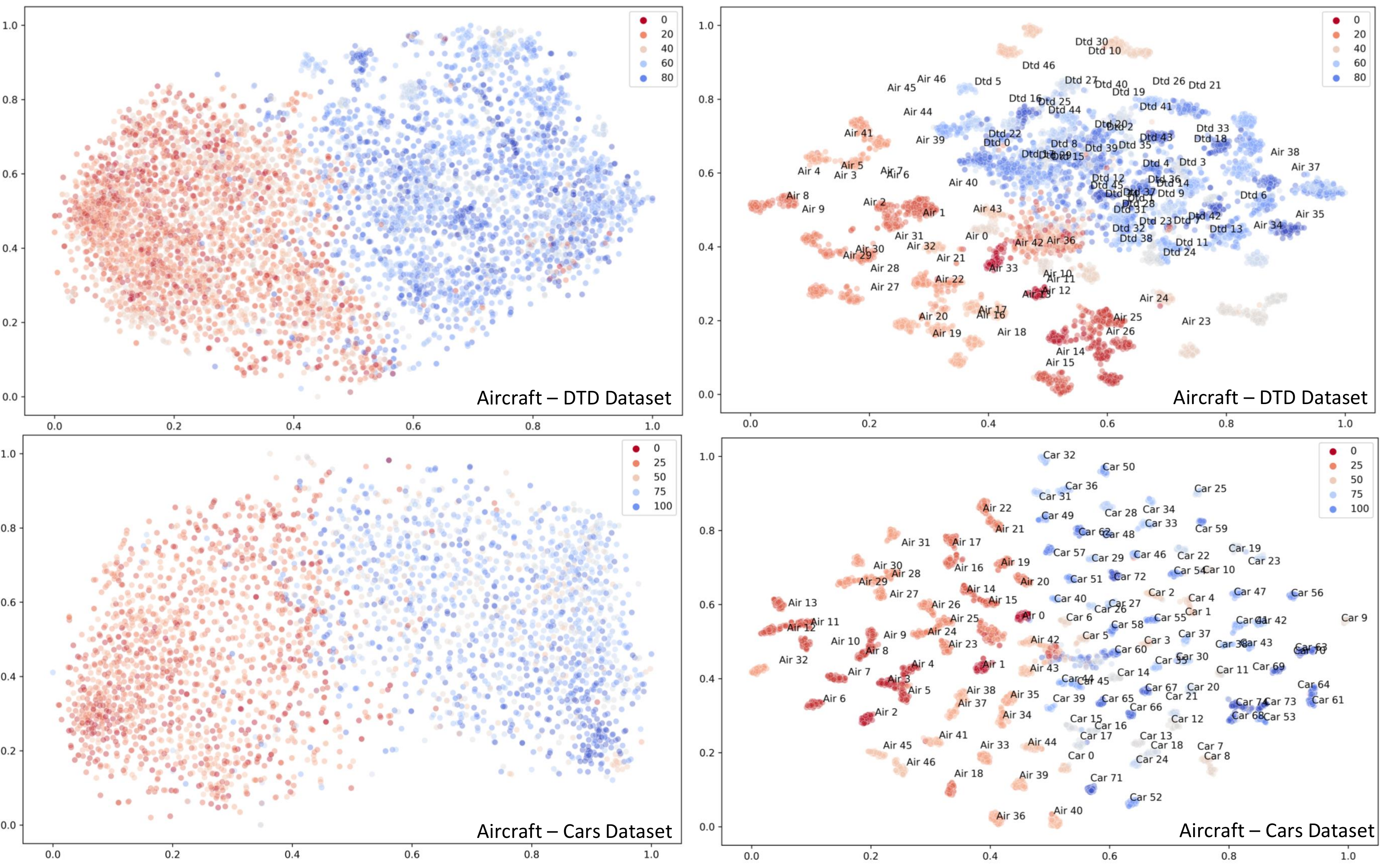}
  \caption{\textbf{T-SNE~\cite{hinton2002stochastic} visualisation of Features Distribution. }Top left: features are derived from ImageNet-1K pre-trained AMF; Top right: features are derived from fine-tuned AMF in Aircraft-DTD, labels of each cluster are annotated; Bottom left: features are derived from ImageNet-1K pre-trained AMF; Bottom right: features are derived from fine-tuned AMF in Aircraft-Cars, labels of each cluster are annotated. Colors are scaled by the class number: $0-46$ in red refers to samples in Aircraft-47; $47-93$ in blue refers to samples in DTD-47; $47-121$ in blue refers to samples in Cars-75. }
  \label{fig:fa}
\end{figure*}

Figure~\ref{fig:fa} plots the T-SNE visualisation of feature distributions derived by Adaptable Multi-tuning Networks. The distribution scatters indicate that pre-trained weights of the model cannot separate aircraft's features from DTD's features well prior to fine-tuning. After fine-tuning with Adaptable Multi-tuning Networks on Aircraft-DTD and Aircraft-Cars, the distribution of each class are clustered together, and distribution between different classes are distinctly separated, resulting in high classification accuracy. The feature analysis supports that Adaptable Multi-tuning Networks can converge and have a high performance on Aircraft-DTD and Aircraft-Cars.

\subsubsection{Stability of AMF\\} 

From empirical results, small learning rates contribute positively to raising the stability. We also control the initialization of the weights of full-connected layers, sampled from a normal distribution with $0$ mean and $0.1$ standard deviation.

\subsubsection{Strategies to choose backbones of policy networks\\}

Backbones of policy networks depend on the distribution of target datasets. In above experiments, $ResNet-34$ has enough ability and complexity to overcome the mixture distribution. We suggest a shallow model is preferred to avoid over-fitting, and reduce computational resources. The use of different policy networks is also a potential extension in the future. 

\subsubsection{Different decay epochs in test stage and validation stage\\}
A decaying learning rate every $20$ epochs contributes more benefit to validation accuracy. In both Aircraft-DTD and Aircraft-Cars, the training set and validation set of all datasets are equivalently separated. When the size of the training data doubles, the numbers of batches in each epoch also double, decelerating the decline in learning rate. Therefore, a smaller decay epoch is required when training with the training and the validation set combined for prediction on the test set.  Decay epochs are decreased to $11$ and $14$ for Aircraft-DTD and Aircraft-Cars respectively in the test stage. 

The stability of the policy network is also closely related to learning rates decay from empirical results. For instance, when a sub-optimal decay is set, one confidence score will be stuck at $0.99$. Therefore, proper adjustment of decay based on the amount of training data is necessary for Adaptable Multi-tuning.

\subsubsection{Dynamic Changes of Weighting Vector}

\begin{figure}[t]

\begin{minipage}{0.45\textwidth}
\includegraphics[width=\textwidth]{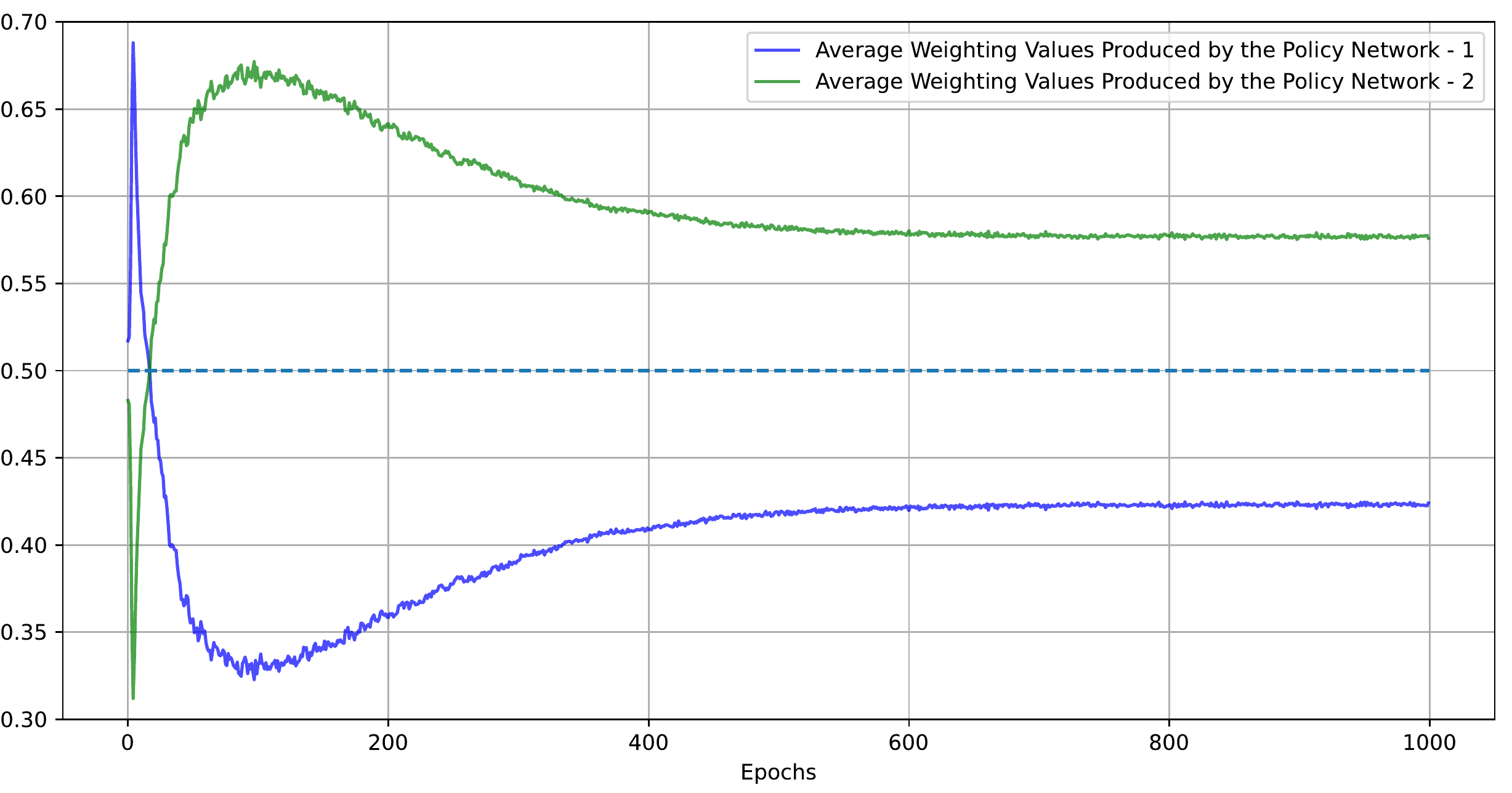}
\end{minipage}
\begin{minipage}{0.45\textwidth}
\includegraphics[width=\textwidth]{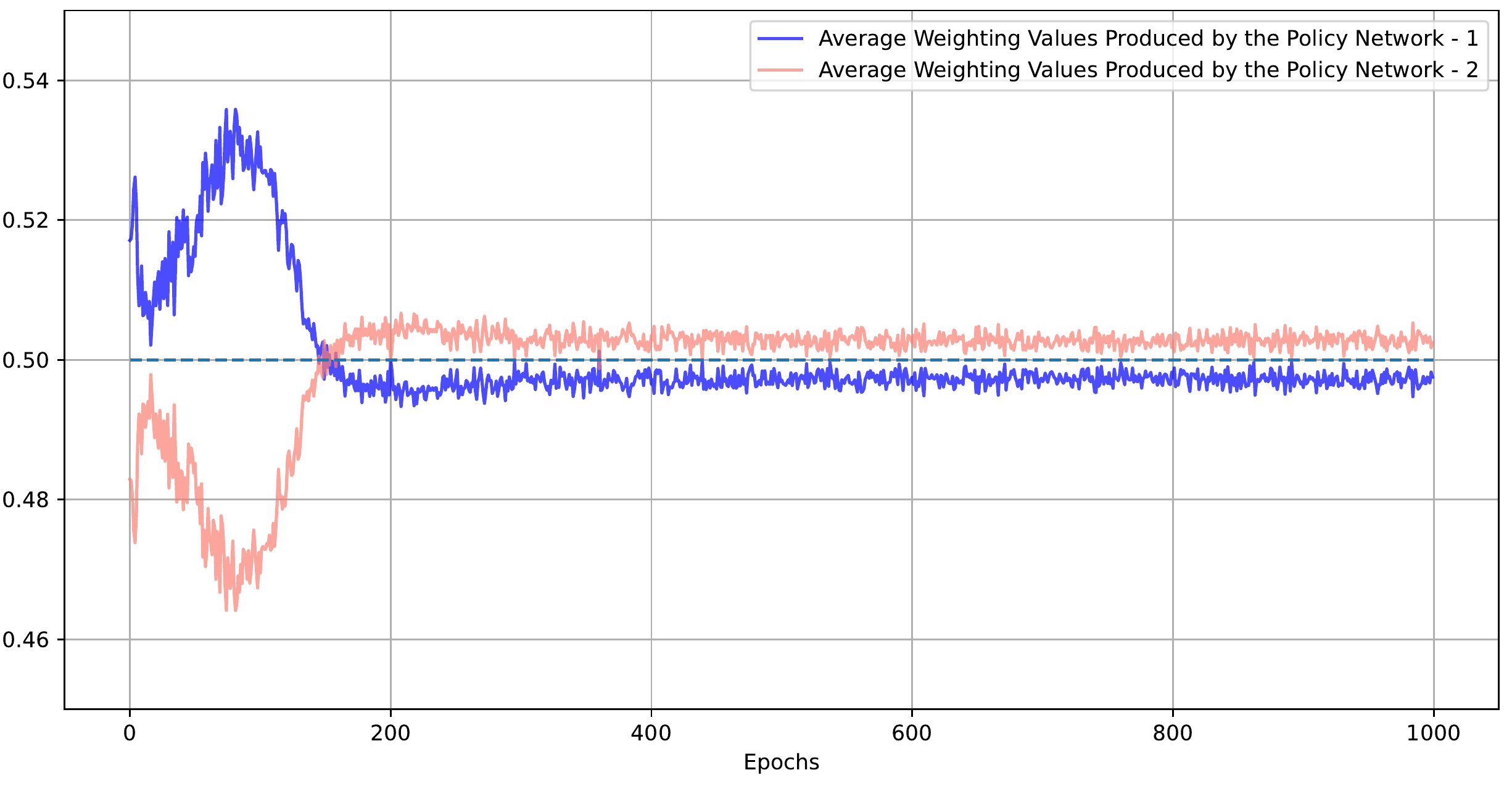}
\end{minipage}

\caption{\textbf{Dynamic Changes of Averaged Weighting Values} during training on Aircraft-DTD (Left) and Aircraft-Cars (Right). We only report the monitor at first $1000$ epochs, due to the page width.}
\label{fig:ecsad}
\end{figure}

We monitor the dynamic changes of weighting vectors, produced by the policy network; and plot in figure~\ref{fig:ecsad}. Weighting vectors aim to scale the feature information for proper feature fusion. 

The averaged values start at around $0.5$, as expected. It reaches extreme values $0.35$ in Aircraft-DTD and $0.46$ in Aircraft-Cars, respectively. In Aircraft-DTD, the policy network converges around $600$ epochs; while the value becomes stable around $200$ epochs in Aircraft-Cars. 

Unexpectedly, the relative positions of two weighting values are flipped during the training on both Aircraft-DTD and Aircraft-Cars. 

\subsubsection{Assignment Accuracy}

\begin{figure}[t]

\begin{minipage}{0.45\textwidth}
\includegraphics[width=\textwidth]{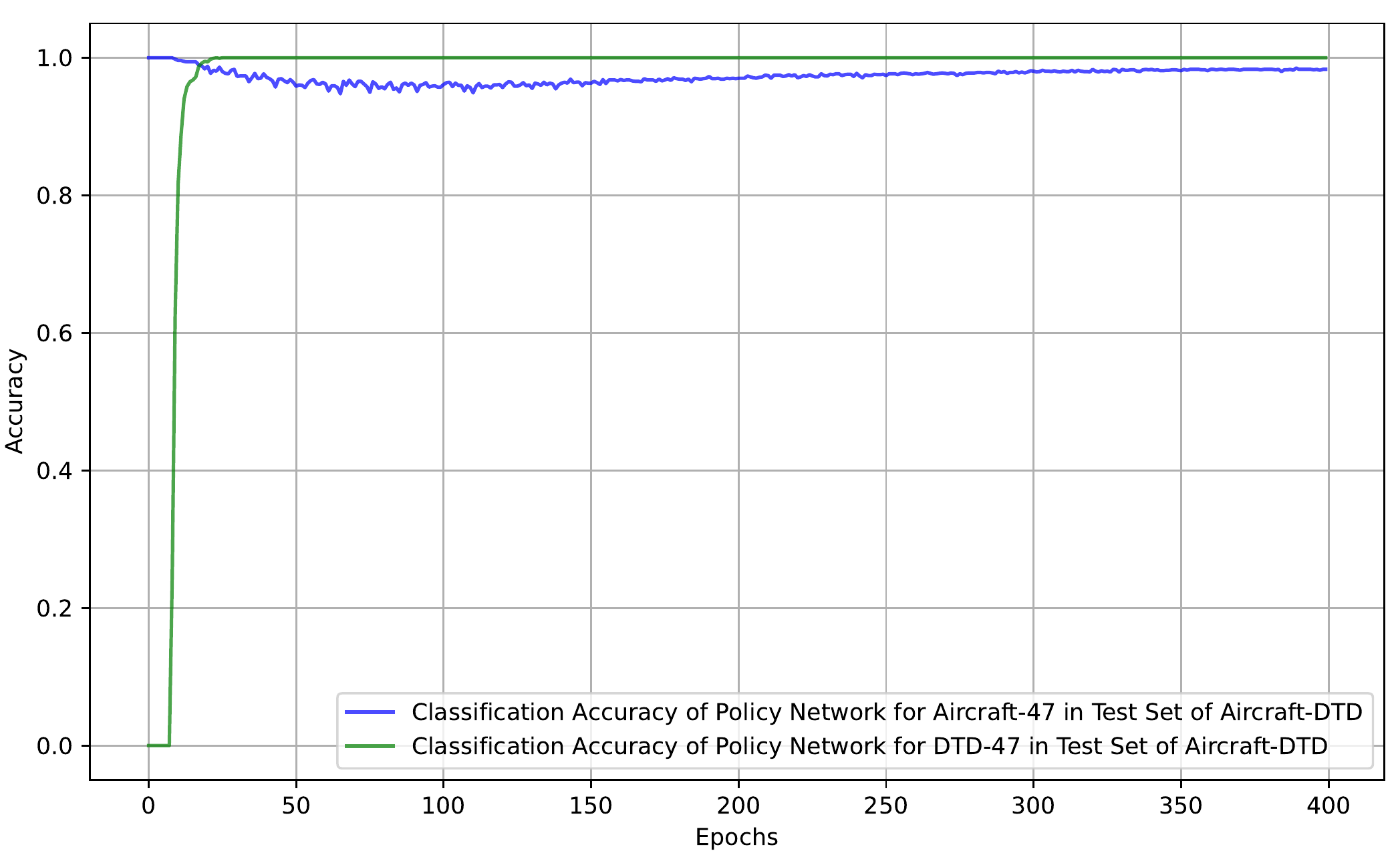}
\end{minipage}
\begin{minipage}{0.45\textwidth}
\includegraphics[width=\textwidth]{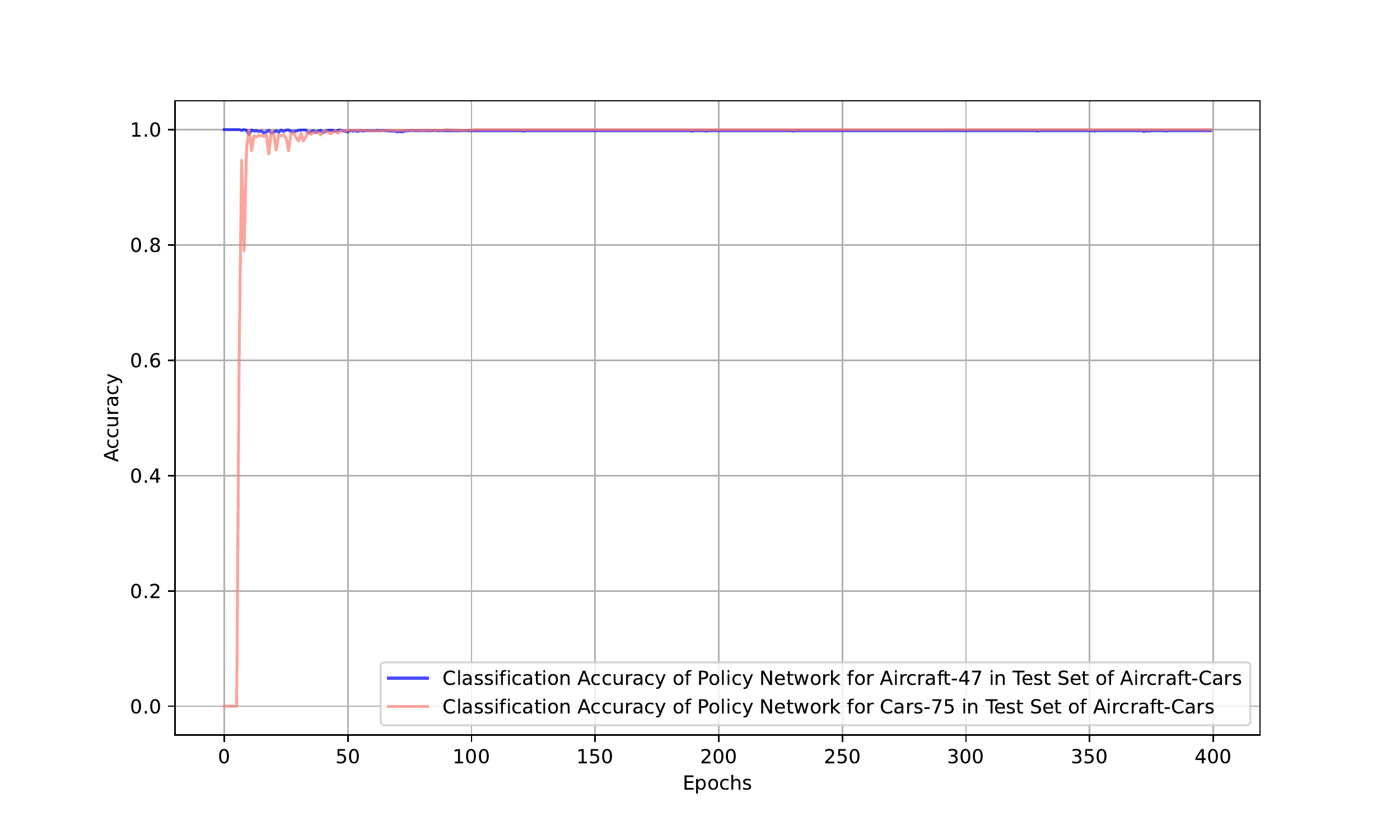}
\end{minipage}

\caption{\textbf{Assignment Accuracy from Policy Network} during training on Aircraft-DTD (Left) and Aircraft-Cars (Right). The accuracy of each class is calculated from: \\            
$\frac{number \ of \ correct  \ assignment}{total \ number \ of \ samples \ in \ the \ class}$. Not all experimental results are plotted in order to enlarge the critical ranges and minimize the converged range.}
\label{fig:monitor}
\end{figure}

We also implement a dynamic monitor (in Figure~\ref{fig:monitor}) to record changes in the accuracy of assignment from the policy networks throughout training. 

The policy network always predicts an embedding space that assigns all samples into the one class at the initial stage (first $20$ epochs). In other words, the assignment accuracy is $50\%$. The Network rapidly adjust its prediction through the plenty from the gradient during the first $50$ epochs in both Aircraft-DTD and Aircraft-Cars. Then, it converges after the first $150$ epochs, where accuracy fluctuation are reduced. 

Evetually, in the test-set of Aircraft-DTD, the trained policy network only miss-assigns $0.77\%$ mixed samples of Aircraft into DTD; all DTD samples are correctly assigned. The policy network behaves $100\%$ accuracy assignment in the test-set of Aircraft-Cars.

\section{Conclusion}

We propose a novel adaptive multi-tuning framework named Adaptable Multi-tuning in this work. It can control multiple fine-tuning parameters in the prediction network. The policy network in Adaptive Multi-tuning Framework adaptably weights the contributions of each fine-tuned model to the final classifier. Our method is also fully differentiable, and so can be implemented under any modern deep learning approach.

In our experiments, two mixture distributions are introduced for evaluating the performance of different methods, using Aircraft-DTD and Aircraft-Cars. The empirical results indicate that standard fine-tuning performs well on single data distributions and simple mixture distributions such as Aircraft-Cars, but behaves with limited performance in complex distributions Aircraft-DTD. Adaptable Multi-tuning Framework is shown to break the limited performance of standard fine-tuning and achieve comparable accuracy in all datasets used. Compared to traditional Multi-Tune, AMF have $4.46\%$ test accuracy improvement at maximum. Our method outperforms the state of the art single model, which is a surprising result given the complexities of prediction with two datasets combined. We also show that Adaptable Multi-tuning is an efficient algorithm, by clarifying the mechanism of the policy network and latent space analysis. Several suggestions related to the use of our method are provided, including hyper-parameters and backbones. Importantly, we illustrate that optimal learning rates for standard fine-tuning is also highly possible to be optimal for sub-networks in Adaptable Multi-tuning Network.

Adaptable Multi-tuning Framework is a novel approach in transfer learning, and also it firstly integrates neural architecture search with fine-tuning. There are several extensible works in the future. Testing Adaptable Multi-tuning in more scenarios where mixture distributions occur, such as face recognition, is meaningful and practicable.  In this paper, we only attempt Inception-v4 and ResNet-34 for Adaptable Multi-tuning Framework; other CNN-based and transformer-based models are expected to tested in our following works.

\bibliographystyle{IEEEtran}
\bibliography{egbib}

\end{document}